\pdfoutput=1
\documentclass[11pt]{article}

\usepackage{EMNLP2022}
\usepackage{times}
\usepackage{latexsym}

\usepackage[T1]{fontenc}
\usepackage[utf8]{inputenc}

\usepackage{microtype}

\usepackage{inconsolata}

\usepackage{graphicx}
\usepackage{xspace}
\usepackage{CJKutf8}
\usepackage{booktabs}
\usepackage{multirow}
\usepackage{diagbox}
\usepackage{color}
\usepackage{xcolor}
\usepackage{soul}
\usepackage{lscape}
\usepackage{amsmath}
\usepackage{siunitx}
\usepackage{dcolumn}
\usepackage{pifont}
\usepackage{tikz}
\usepackage{makecell}
\usepackage{array}
\usepackage{amssymb}
\usepackage{enumitem}

\definecolor{shadecolor}{rgb}{0.92,0.92,0.92}
\definecolor{lightblue}{RGB}{223, 233, 248}
\definecolor{lightblue1}{HTML}{C5DAFB}
\sethlcolor{lightblue1}

\newcommand{\UnifiedABSA}{\textsc{UnifiedABSA}\xspace}
\newcommand{\USI}{\textsc{USI}\xspace}
\newcommand{\template}[1]{{\emph{#1}}}
\newcommand{\single}{\textit{Single}\xspace}

\title{\UnifiedABSA: A Unified ABSA Framework Based on Multi-task Instruction Tuning}

\author{Zengzhi Wang, Rui Xia\thanks{\, Corresponding author} \and Jianfei Yu \\
       School of Computer Science and Engineering, \\ Nanjing University of Science and Technology, China \\
       \texttt{\{zzwang, rxia, jfyu\}@njust.edu.cn}
       }

\begin{document}
\maketitle
\begin{abstract}

Aspect-Based Sentiment Analysis (ABSA) aims to provide fine-grained aspect-level sentiment information. There are many ABSA tasks, and the current dominant paradigm is to train task-specific models for each task. However, application scenarios of ABSA tasks are often diverse. This solution usually requires a large amount of labeled data from each task to perform excellently. These dedicated models are separately trained and separately predicted, ignoring the relationship between tasks. 
To tackle these issues, we present \UnifiedABSA, a general-purpose ABSA framework based on multi-task instruction tuning, which can uniformly model various tasks and capture the inter-task dependency with multi-task learning. 
Extensive experiments on two benchmark datasets show that \UnifiedABSA 
can significantly outperform dedicated models on 11 ABSA tasks and show its superiority in terms of data efficiency.

\end{abstract}

\section{Introduction}

\begin{figure}[t]
\centering 
\includegraphics[scale=0.34]{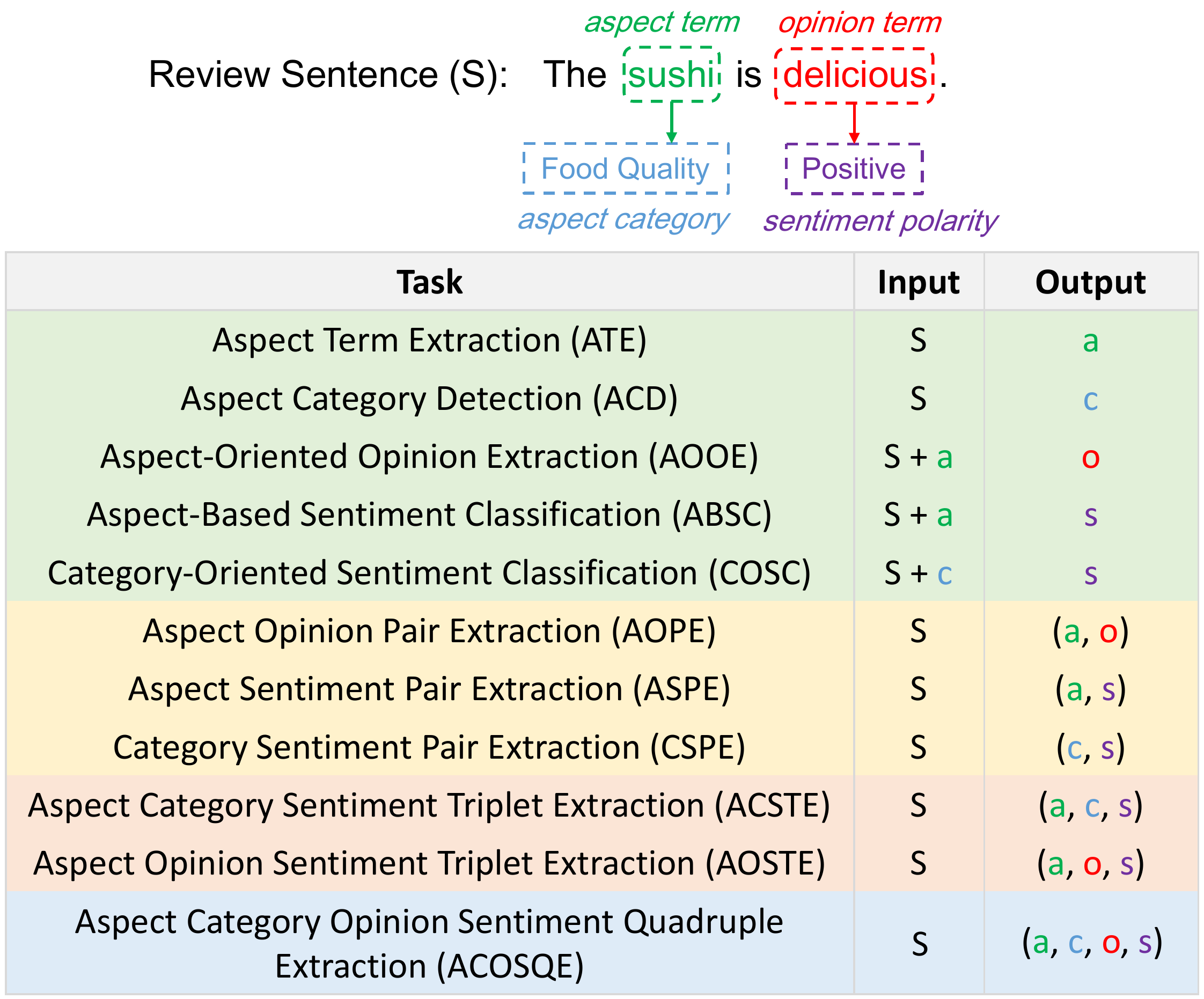} 
\caption{An example of various ABSA tasks. \emph{a}, \emph{c}, \emph{o}, and \emph{s} stand for the aspct term, aspect category, opinion term, and sentiment, respectively.} 
\label{fig:various-absa-tasks-example}
\end{figure}

Aspect-Based Sentiment Analysis (ABSA) has received much attention due to its wide application. It aims to mine fine-grained aspects of reviewed entities and determine the sentiment for each aspect \citep{liu2012sentiment,pontiki-etal-2014-semeval}. Generally, there are four key elements \emph{aspect term (a)}, \emph{aspect category (c)}, \emph{opinion term (o)}, and \emph{sentiment polarity (s)} in ABSA:
\emph{aspect term} is the opinion target in terms of an entity and its aspect; \emph{opinion term} is a subjective statement on the aspect; \emph{aspect category} denotes a unique predefined category of the aspects; \emph{sentiment polarity} denotes the predefined semantic orientation (e.g., Positive, Negative, or Neutral).
\emph{aspect term} and \emph{opinion term} are usually a word or phrase in the text; \emph{aspect category} and \emph{sentiment polarity} are predefined class labels.
As shown in Figure~\ref{fig:various-absa-tasks-example}, given a review sentence, ``\emph{The sushi is delicious.}'', the corresponding four elements are ``\emph{sushi}'', Food Quality, ``\emph{delicious}'', and Positive.

Grouping different types of ABSA elements yields a variety of ABSA tasks, including ATE, ACD, AOOE, ABSC, COSC, AOPE, ASPE, CSPE, ACSTE, ACOSQE, etc\footnote{Some tasks may be referred to as other appellations in prior works, e.g., AOOE is also known as Target-oriented Opinion Words Extraction (TOWE). For the consistency in terminology, we refer to it as AOOE in this work. The same goes for other tasks.}.
Figure~\ref{fig:various-absa-tasks-example} illustrates eleven representative ABSA tasks.

In the literature, all the aforementioned ABSA tasks have been explored to tackle different downstream applications.
For example, ABSC aims to identify the sentiment polarity for the given aspect term~\cite{tang2016effective}, whereas AOPE aims to extract the aspect terms and their associated opinion terms~\cite{chen-etal-2020-synchronous}.
For the most complex task ACOSQE~\cite{cai-etal-2021-acos}, its goal is to extract all the aspect terms with their corresponding aspect categories, opinion terms, and sentiment polarities simultaneously. 
Since the goals of these ABSA tasks are different from each other, existing studies typically train separate task-specific systems for different downstream applications.

However, these task-specific solutions still suffer from several shortcomings as follows: (1) the task-specific models often require a large amount of annotated data for each task, which is time-consuming and laborious; 
(2) these task-specific models are trained separately and predicted separately, failing to exploit the useful knowledge shared in other tasks; 
(3) all the existing ABSA tasks are essentially the combinations of the subset of four sentiment elements, and thus are highly related and complementary,
as observed in prior works~\citep{he-etal-2019-imn,chen-qian-2020-relation,yu-etal-2021-making}. Nevertheless, training dedicated models for each task ignores the inter-task dependency.

To address the above limitations, we present \UnifiedABSA, a general-purpose ABSA framework based on multi-task instruction tuning that can uniformly model various ABSA tasks and capture the inter-task dependency with multi-task learning. 
Specifically, we formulate all ABSA tasks as a conditional
generation problem~\citep{yan-etal-2021-unified,zhang-etal-2021-gas,zhang-etal-2021-paraphrase} based on the latest text-to-text pre-training language models (PLMs), such as T5~\citep{raffel2020t5} and BART~\citep{lewis-etal-2020-bart}. To distinguish between various ABSA tasks, we design unified sentiment instructions (\USI) for each task to prompt the model. Conditioned on a review sentence wrapped in \USI, the model learns to paraphrase the sentence into a natural summary sentence following a predefined task-specific template in \USI.

Compared with existing task-specific and multi-task learning (MTL) models, our \UnifiedABSA framework shows its superiority in data efficiency.
First, instead of utilizing data from different tasks and sources in existing task-specific models and MTL studies~\citep{khashabi-etal-2020-unifiedqa,DBLP:journals/corr/abs-2201-05966-unifiedskg}, \UnifiedABSA only requires the annotated data from the ACOSQE task to derive the annotation for all the eleven ABSA tasks. 
Second, since our designed \USI provides much more semantic information, which can instruct the pre-trained language model to learn better under the low-resource setting, \UnifiedABSA is expected to perform well when only a small amount of training data are available.

Experimental results on two ACOS datasets show that \UnifiedABSA yields highly competitive performance on 11 tasks and boosts average performance by 1-2\% in the full-supervised setting (\S~\ref{sec:full-data-exp}). In an extremely low resource setting (only 32 examples), \UnifiedABSA can outperform task-specific models by 5\% F1 score in average performance on 11 ABSA tasks (\S~\ref{sec:low-resource-exp}).
More importantly, we also find that our \UnifiedABSA can achieve roughly the same performance with half of the data size as dedicated models in the low-resource scenario (less than 512 samples). These results demonstrate the effectiveness of \UnifiedABSA and its advantage in terms of data efficiency.

\section{Related Work}

\noindent\textbf{Aspect-Based Sentiment Analysis} \quad As a fine-grained sentiment analysis task, it has received much attention. Early studies mainly focus on Aspect-Based Sentiment Classification (ABSC)~\citep{dong-etal-2014-adarnn,wang-etal-2016-attention,tang2016effective,xue-li-2018-aspect}. Later, researchers attempt to jointly extract aspects and predict the corresponding sentiment, known as End-to-End ABSA~\citep{he-etal-2019-imn,DBLP:conf/aaai/LiBLL19,hu-etal-2019-open,chen-qian-2020-relation}. A similar development trend can also be observed from Aspect Opinion Co-Extraction (AOCE)~\citep{liu-etal-2015-fine,wang-etal-2016-recursive,DBLP:conf/aaai/WangPDX17}, Aspect-Oriented Opinion Extraction (AOOE)~\citep{fan-etal-2019-target,wu2020latent} to Aspect-Opinion Pair Extraction~\citep{chen-etal-2020-synchronous,zhao-etal-2020-spanmlt}. Recently, more complex ABSA tasks, i.e., Aspect-Opinion-Sentiment Triplet Extraction (AOSTE)~\citep{peng2020knowing,xu-etal-2020-position,wu2020grid,mao2021dualmrc,yan-etal-2021-unified,xu-etal-2021-learning,zhang-etal-2021-gas}, Aspect-Category-Sentiment Triplet Extraction (ACSTE)~\citep{wan2020tas}, and Aspect-Category-Opinion-Sentiment Quad Extraction (ACOSQE)~\citep{cai-etal-2021-acos,zhang-etal-2021-paraphrase} have been proposed to provide complete fine-grained sentiment information. However, existing studies mainly focus on building task-specific dedicated models. Recently, \citet{yan-etal-2021-unified} attempted to explore a unified framework for various ABSA tasks to avoid excessive task-specific architecture designs. Despite its effectiveness, it still trains task-specific models for each task. Compared with previous studies, our work aims to build a general-purpose multi-task ABSA framework that uniformly model various ABSA tasks, performs multi-task learning, and takes full advantage of the relationship between tasks.

\noindent\textbf{Prompt (Instruction) Learning} \quad Prompt learning aims to bridge the gap between pre-training objectives and downstream tasks~\citep{liu2021prompt-survey}. Researchers have designed various natural language templates (known as \emph{discrete prompt}) to wrap the original input and thus prompt the PLMs. This practice has shown great success in a variety of NLP tasks~\citep{DBLP:conf/nips/Brown20-GPT3,schick-schutze-2021-few-shot-generation,schick-schutze-2021-just,gao-etal-2021lm-bff,DBLP:journals/corr/abs-2109-04645-CINS,seoh-etal-2021-open}. In this work, we also adopt a similar operation (i.e., wrap the original input) to prompt the PLM. A comprehensive overview of prompt learning can refer to~\citet{liu2021prompt-survey}. \textbf{Instruction learning} aims to teach the model to follow language instructions like humans to perform many new tasks~\cite{DBLP:journals/corr/abs-2010-11982-turing-test,mishra-etal-2022-natural-instruction}. Instruction learning is essentially the same as prompt learning, which induces language models to perform the specific end task better. The method we adopt is collectively called \emph{multi-task instruction tuning} for consistency. It can also be called \emph{multi-task prompt training}.

\noindent\textbf{Multi-task Learning} \quad It is a promising direction to build a general-purpose model that performs a wide range of NLP tasks to share the general language knowledge~\citep{DBLP:journals/ml/Caruana97,DBLP:conf/icml/CollobertW08,DBLP:journals/corr/abs-1806-08730-decathlon}. Earlier, T5~\citep{raffel2020t5} converted all NLP tasks into the text-to-text format and performed multi-task learning via the task prefix. However, it is inferior to its corresponding dedicated model. Recently, some models empowered by the pre-training language models (PLMs), such as T5~\citep{raffel2020t5} and BART~\citep{lewis-etal-2020-bart}, have shown promising multi-task learning results on several language tasks, e.g., Question Answering (QA)~\citep{khashabi-etal-2020-unifiedqa,zhong-etal-2022-proqa}, Structured Prediction (SP)~\citep{DBLP:conf/iclr/Paolini21tanl}, Information Extraction (IE)~\citep{lu-etal-2022-uie} and Structured Knowledge Grounding (SKG)~\citep{DBLP:journals/corr/abs-2201-05966-unifiedskg}. However, there is no similar study in the ABSA community to build a general ABSA system. More recently, FLAN~\citep{wei2021flan} and T0~\citep{sanh2021t0}, built on the latest PLMs, have shown promising ability on cross-task generalization via instruction learning~\citep{mishra-etal-2022-natural-instruction} or prompt learning~\citep{liu2021prompt-survey}. Inspired by these, this work adopts multi-task instruction tuning~\citep{wei2021flan} to build a general ABSA framework, focusing on multi-task learning rather than cross-task generalization.

\begin{figure*}[!h]
\centering 
\includegraphics[width=2.0\columnwidth]{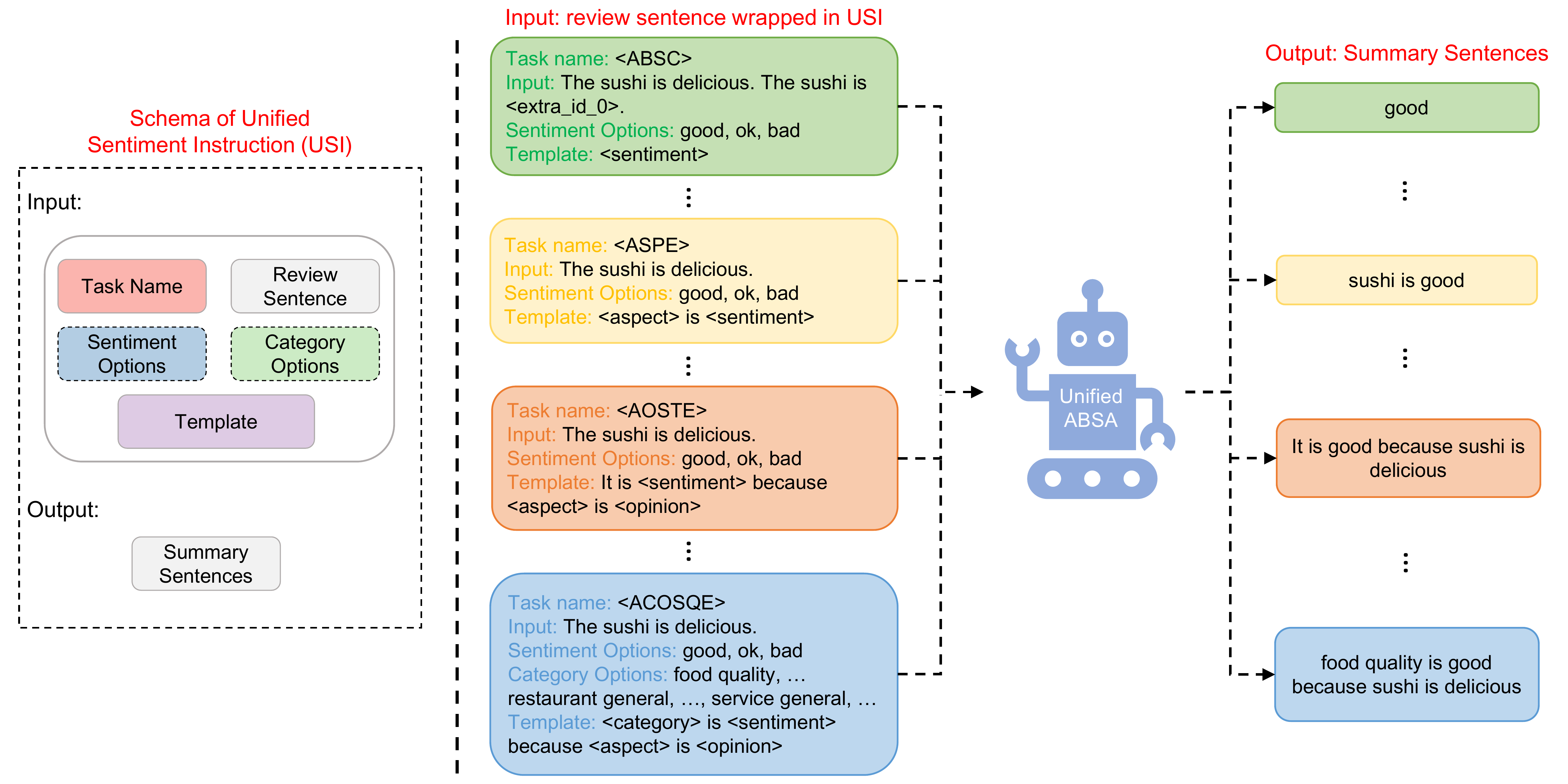} 
\caption{Left: An overview of \USI wrapped input and output. The input consists of the original review sentence, task name, sentiment options, category options, and template, where sentiment options and category options are optional and not required for some tasks (e.g., AOPE). Right: The overview framework of \UnifiedABSA. A few task instances are shown here.} 
\vspace{-3mm}
\label{fig:Unified-Model}
\end{figure*}

\section{\UnifiedABSA}

Generally, \UnifiedABSA aims to uniformly model the text-to-text formulation of various ABSA tasks based on a single framework, where corresponding labels are paraphrased as summary sentences in natural language. As shown in Figure~\ref{fig:Unified-Model}, given a review sentence and our designed \emph{unified sentiment instruction} (\USI) (\S~\ref{sec:usi}), the model is asked to generate a paraphrased summary output following the corresponding task-specific template~(\S~\ref{sec:summary-sentence-generation}). We take an off-the-shelf text-to-text PLM (i.e., T5) and fine-tune it on a collection of 11 ABSA tasks verbalized via unified sentiment instructions (\USI) in natural language. We refer to it as \emph{multi-task instruction tuning} (\S~\ref{sec:multi-task-instuction-tuning}).

\subsection{Unified Sentiment Instruction}
\label{sec:usi}

We design unified sentiment instruction to teach the language model to perform various ABSA tasks. These instructions can fully elicit the capability of language models, help the model recognize different tasks, and teach the model to perform ABSA tasks excellently.

\subsubsection{The Schema of \USI}

Conditioned on the \USI wrapped original review sentence shown in Figure~\ref{fig:Unified-Model}, our \UnifiedABSA model is taught to generate the corresponding paraphrase summary output. The design of \USI follows a unified schema~(on the left side of Figure ~\ref{fig:Unified-Model}). \USI consists of the \emph{task name}, \emph{sentiment options}, \emph{category options}, and \emph{template}. Each element of \USI under the schema is described as follows:

\noindent\textbf{Task Name}: We design it for each task to help the model distinguish different ABSA tasks.

\noindent\textbf{Sentiment Options}: We define three sentiment label words, \emph{good}, \emph{ok}, and \emph{bad}, to represent \emph{positive}, \emph{neutral}, and \emph{negative} sentiment polarities. It makes the model aware of which sentiment choices are desired when it comes to those tasks related to sentiment classification, such as ABSC, ASPE, AOSTE, and ACOSQE.

\noindent\textbf{Category Options}: We consider the related predefined aspect categories as this element. It also makes the model aware of which category choices are desired when it comes to those tasks related to category detection, such as ACD, CSPE, ACSTE, and ACOSQE.

\noindent\textbf{Template}: We design it for each task. It gives the model more fine-grained task-specific guidance, i.e., the model will perform the more controlled condition generation via prompting it with the desired paraphrase template.

We concatenate all the above elements of \USI and the original review sentence with ``\verb|\|n'' as the separator, and each element start with the corresponding leading prefix token (``\template{Task Name:}'', ``\template{Input:}'', ``\template{Sentiment Options:}'', ``\template{Category Options:}'', ``\template{Template:}'').

\subsubsection{USI for ABSA Tasks}
\label{sec:USI-for-ABSA}

\begin{figure}[h]
\centering 
\includegraphics[width=1.0\columnwidth]{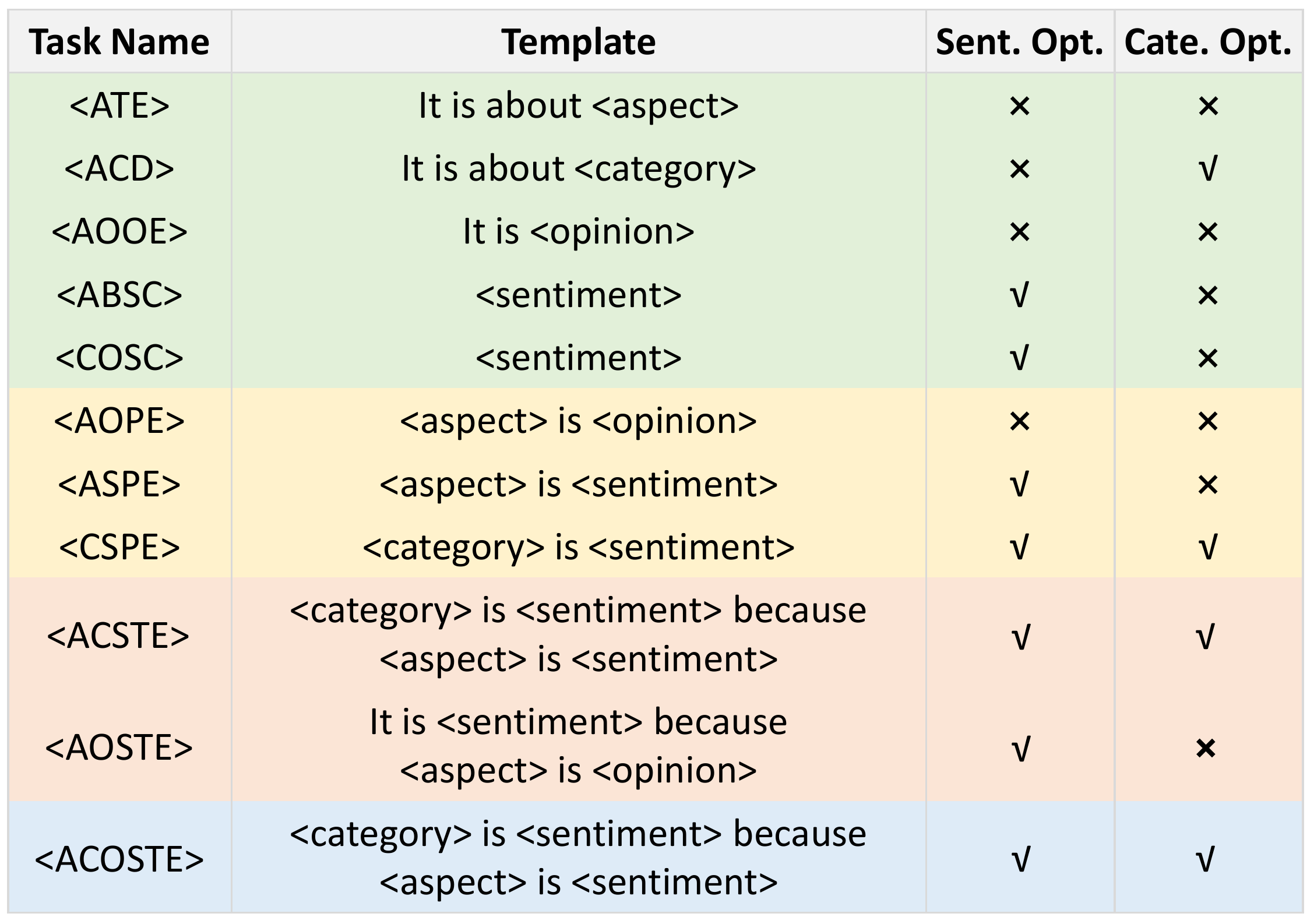} 
\caption{An overview of \USI designed for each task. We define the corresponding task name and template for each task. $< \cdot >$ denotes a special token. \emph{Sent. Opt.} and \emph{Cate. Opt.} stand for \emph{Sentiment Options} and \emph{Category Options}, respectively. The last two columns indicate whether the \USI designed for each task contains these two options.} 
\vspace{-3mm}
\label{fig:template-for-all-tasks}
\end{figure}

Our specific design of \USI for each task is shown in Figure~\ref{fig:template-for-all-tasks}. All ABSA tasks are provided with the \emph{task name} and \emph{template}. Taking the ACOSQE task as an example, task name is ``\emph{<ACOSQE>}'' while ``\emph{<category> is <sentiment> because <aspect> is <opinion>}'' is the corresponding template. For these tasks involving category detection like ACOSQE, we use ``\template{Category Options:}'' as an indicator followed by the relevant candidate categories, separated by the comma. Similarly, we use ``\template{Sentiment Options:}'' as an indicator followed by the sentiment label words for those tasks involving sentiment classification.
Figure~\ref{fig:Unified-Model} shows the four task instances.

Besides, it is worth noting that we make a simple modification to the original review sentence for the task given an aspect, i.e., ABSC COSC and AOOE. Specifically, for the ABSC and COSC tasks, the original review sentence will be followed by ``\emph{The \{aspect\} is <extra\_id\_0>}'', where ``\{aspect\footnote{It will be replaced with the actual aspect category for the COSC task.}\}'' will be replaced by the actual aspect term and ``<extra\_id\_0>'' is a special token during T5 pre-training process. It allows for better utilization of the language model by modeling the downstream task as the pre-training objective (i.e., \emph{mask language modeling}) of T5. This practice is known as \emph{prompt learning}~\citep{liu2021prompt-survey}, which has shown promising results on a myriad of NLP tasks~\citep{schick-schutze-2021-few-shot-generation,schick-schutze-2021-just,gao-etal-2021lm-bff}. For the AOOE task, the original review sentence will be followed by ``\emph{What about the \{aspect\}?}'', where ``\{aspect\}'' will also be replaced as well.

\subsection{Summary Sentence Generation}
\label{sec:summary-sentence-generation}

We paraphrase the label as summary sentence(s) following the task-specific template as introduced in~\S~\ref{sec:USI-for-ABSA}, where the placeholders \emph{<aspect>}, \emph{<category>}, \emph{<opinion>}, and \emph{<sentiment>} will be replaced with actual terms. Given a review sentence $s$ and \USI $u$ as the input, \UnifiedABSA will adaptively generate corresponding sentences. We separate them with a special symbol \texttt{[SSEP]} when there is more than one summary sentence for a given input.

\subsection{Multi-Task Instruction Tuning}
\label{sec:multi-task-instuction-tuning}

With the text-to-text paradigm, all inputs (\S~\ref{sec:usi}) and outputs (\S~\ref{sec:summary-sentence-generation}) of all ABSA tasks are converted into plain text format. To perform the multi-task learning, we directly combine all training instances described via natural language instructions and templates from all ABSA tasks. 

\vspace{-0.1cm}
\begin{equation}
    \mathcal{D} = \bigcup_{i=1}^{k} \bigcup_{j=1}^{|T_i|} \text{instance}_j
\end{equation}
where $k$ is the number of tasks, and $T_i$ denotes the dataset corresponding to the $i$ th task. We fine-tune an off-the-shelf T5 on the mixture $\mathcal{D}$ by minimizing the negative log-likelihood loss using teacher-forcing~\citep{DBLP:journals/neco/WilliamsZ89}.  We refer to the resulting model as \UnifiedABSA.

By default, each training batch randomly samples from the mixture. This similar operation has also been adopted by \citet{raffel2020t5} and \citet{khashabi-etal-2020-unifiedqa}. Subsequent experimental results prove that this is a simple and effective approach. Another alternative operation (uniformly sampling training data for each task within a batch) is also attempted and we will discuss them in \S~\ref{sec:ablation-on-sampling}. More importantly, we highlight that it can be easily extended to new tasks without model architecture modifications.

\section{Experiments}

We conduct extensive experiments on different ABSA tasks and data settings to verify the effectiveness of \UnifiedABSA.

\subsection{Experimental Setup}
\label{sec:exp-setup}

\noindent\textbf{Datasets}. We consider two datasets, Restaurant-ACOS~\citep{cai-etal-2021-acos} and Laptop-ACOS~\citep{cai-etal-2021-acos}, which are two relatively large ACOS datasets and can support various ABSA tasks from simple to complex. Table~\ref{tab:acos-data-statistics} shows the statistics.

\begin{table}[htbp]
\centering
\scalebox{0.90}{
\begin{tabular}{c|c|c}
\toprule
     & \textbf{Restaurant-ACOS} & \textbf{Laptop-ACOS} \\
\midrule
\#Cate. &  \makebox[3ex][r]{13} & \makebox[3ex][r]{121} \\
\#Quad. & \makebox[3ex][r]{3658} & \makebox[3ex][r]{5758} \\
\midrule
\#Train  & \makebox[3ex][r]{1531} & \makebox[3ex][r]{2934} \\
\#Dev  & \makebox[3ex][r]{170}  & \makebox[3ex][r]{326} \\
\#Test & \makebox[3ex][r]{585} & \makebox[3ex][r]{816} \\
\bottomrule
\end{tabular}}
\caption{Statistics of Restaurant-ACOS and Laptop-ACOS deriving from~\citet{cai-etal-2021-acos}. \#Cate. and \#Quad. denote the number of unique categories and ACOS quadruples contained in the dataset, respectively.}
\label{tab:acos-data-statistics}  
\end{table}

\noindent\textbf{Annotated Data Conversion to Other Tasks}. As mentioned earlier, \UnifiedABSA only needs the quadruple data instead of the data from different tasks and sources. Then the data for other subtasks could derive from the ACOS quadruple data. Taking the AOPE task as an example, the corresponding label is $set\left( \{\left(a_i,o_i\right) \} \right)$\footnote{\emph{set($\cdot$)} denotes the de-duplication operation.} for the given review sentence and the ACOS label $\{\left(a_i, c_i, o_i, s_i\right)\}$. The practice allows us to utilize the ACOS data fully. More details are given in Appendix~\ref{sec-appendix:acos-data-conversion}.

\noindent\textbf{Task-Specific Baselines}. Prior work~\cite{zhang-etal-2021-gas,zhang-etal-2021-paraphrase} based on the text-to-text paradigm has successively shown state-of-the-art performance on several ABSA tasks successively. We choose the text-to-text-based approach \emph{Paraphrase}~\cite{zhang-etal-2021-paraphrase} to train dedicated models (also with \textsc{T5-Base}) for each task and  collectively refer to these models as \textbf{\single}. Specifically, we adopt the same output template as \emph{Paraphrase} and use a template of our design (see \S~\ref{sec:USI-for-ABSA}) for those tasks it did not cover. The \single model generates paraphrase sentences following the task-specific template conditioned on the original review sentence.

\noindent\textbf{Evaluation Metric}. We employ micro F1 scores as the evaluation metric. Since the output is in natural language form, we parse it according to the predefined template (see \S~\ref{sec:USI-for-ABSA}) to get the corresponding sentiment tuple. A sentiment tuple is regarded as correct if and only if all sentiment elements inside it are exactly the same as the corresponding gold label.

\noindent\textbf{Implementation Details}. For all baselines and \UnifiedABSA, we fine-tune the pre-trained \textsc{T5-Base}~\citep{raffel2020t5} using the Huggingface Transformers~\citep{wolf-etal-2020-transformers}. We run all experiments with three different random seeds to reduce the variance based on the NVIDIA RTX 3090. We use greedy search for decoding by default and employ the AdamW optimizer~\citep{DBLP:conf/iclr/LoshchilovH19} as the optimizer. We only tune the learning rate within $\{1e$-$4$$, 3e$-$4$$, 5e$-$4\}$ and batch size within $\{4,6,8,12,16,20,24\}$.

\subsection{Experiments on Full-supervised Setting}
\label{sec:full-data-exp}

\begin{table}[]
\centering
\scalebox{0.80}{
\begin{tabular}{l|cc|cc}
\toprule
\multirow{2}{*}{\textbf{Task}} & \multicolumn{2}{c|}{\textbf{Restaurant-ACOS}} & \multicolumn{2}{c}{\textbf{Laptop-ACOS}} \\ \cline{2-5} 
                  & \textbf{Single}        & \textbf{Unify}       & \textbf{Single}     & \textbf{Unify}     \\ \midrule[0.6pt]
                  
ATE      & 84.00   \tiny{0.15}        & \textbf{86.12}  \tiny{0.52}       & 83.57           \tiny{0.15}   & \textbf{85.10}      \tiny{0.20}    \\
ACD   & 83.04   \tiny{0.36 }       & \textbf{84.62}  \tiny{0.09 }      & 59.27           \tiny{0.92  } & \textbf{59.82}      \tiny{0.22 }  \\
ABSC   & 90.66   \tiny{0.33 }       & \textbf{91.95}  \tiny{0.33}       & 89.97           \tiny{0.29}   & \textbf{90.67}      \tiny{0.06}     \\
COSC  & 89.69   \tiny{0.23 }       & \textbf{91.17}  \tiny{0.06 }       & 89.70           \tiny{0.62 }  & \textbf{90.34}      \tiny{0.14} \\
AOOE       & 86.81   \tiny{0.48 }       & \textbf{88.24}  \tiny{0.81 }      & 89.14           \tiny{0.48 }  & \textbf{89.41}       \tiny{0.40} \\
\midrule
ASPE     & 76.33   \tiny{0.13}        & \textbf{78.31}  \tiny{0.41 }      & 76.12           \tiny{0.78}   & \textbf{77.53}      \tiny{0.49} \\
AOPE     & 73.27   \tiny{0.90}        & \textbf{77.76} \tiny{0.60}       & 73.63            \tiny{0.95}   & \textbf{77.32}       \tiny{0.69} \\
CSPE   & 75.81   \tiny{0.44}        & \textbf{77.91} \tiny{0.27}       & 52.95            \tiny{0.61}   & \textbf{55.77}      \tiny{0.57} \\
\midrule
AOSTE      & 68.81   \tiny{0.56}        & \textbf{73.06}  \tiny{0.63}       & 72.06           \tiny{0.97}   & \textbf{75.38}      \tiny{0.04}  \\
ACSTE   & 68.22   \tiny{0.67}        & \textbf{70.16}  \tiny{0.55}       & 47.97           \tiny{0.89}   & \textbf{48.29}      \tiny{0.11}  \\
 \midrule
ACOSQE       & 59.38   \tiny{0.37}        & \textbf{60.60}  \tiny{0.83}       & \textbf{44.24} \tiny{0.64}     & 42.58   
\tiny{0.85} 
\\ \midrule
Ave.                  & 77.82    & \textbf{79.99} & 70.78               & \textbf{72.02} \\

 \bottomrule
\end{tabular}}
\caption{The F1 scores (std) comparison of 11 tasks between single models (Single) and \textsc{Unified-ABSA} (Unify) on two quadruple datasets under the fully-supervised setting. We also report the average performance over 11 tasks. The best results are in bold.}
\vspace{-4mm}
\label{tab:full-data-table}
\end{table}

We conduct experiments under the fully-supervised setting, as shown in  Table~\ref{tab:full-data-table}. \UnifiedABSA achieves 2.17\% and 1.24\% improvements over dedicated models on two datasets in terms of average performance over 11 tasks. Moreover, it consistently outperforms them in individual tasks, except on Laptop-ACOS for ACOSQE task (1.66\% performance drop). One likely explanation for that could be that Laptop-ACOS has more aspect categories than Restaurant-ACOS (i.e., 121 vs. 13, as shown in Table~\ref{tab:acos-data-statistics}), which leads to the fact that it may not always gain from other tasks during multi-task instruction tuning.

\subsection{Experiments on Low-resource Setting}
\label{sec:low-resource-exp}

\begin{table*}[]
\centering
\setlength{\belowcaptionskip}{-0.3cm}
\scalebox{0.85}{

\begin{tabular}{l|cccc|cccc}
\toprule
\multirow{3}{*}{\textbf{Task}} & \multicolumn{4}{c|}{\textbf{Restaurant-ACOS}}                                                                                                                                                             & \multicolumn{4}{c}{\textbf{Laptop-ACOS}}                                                                                                                                                                  \\ \cline{2-9} 
                               & \multicolumn{2}{c|}{\textbf{32-Shot}}                                                                          & \multicolumn{2}{c|}{\textbf{64-Shot}}                                                     & \multicolumn{2}{c|}{\textbf{32-Shot}}                                                                          & \multicolumn{2}{c}{\textbf{64-Shot}}                                                      \\  
                               & {\textbf{Single}} & \multicolumn{1}{c|}{\textbf{Unify}}  & {\textbf{Single}} & \multicolumn{1}{c|}{\textbf{Unify}} & {\textbf{Single}} & \multicolumn{1}{c|}{\textbf{Unify}} & {\textbf{Single}} & \multicolumn{1}{c}{\textbf{Unify}}  \\ \midrule

ATE                            & 61.63             \tiny{$\pm$  5.13}            &\multicolumn{1}{c|}{ \textbf{65.36}        \tiny{$\pm$ 2.10}}    & 68.42          \tiny{$\pm$ 1.16}           & \textbf{70.43}        \tiny{$\pm$ 0.49}      & 64.01                 \tiny{$\pm$ 3.91}        & \multicolumn{1}{c|}{\textbf{69.22}      \tiny{$\pm$ 1.14}}            & 70.99           \tiny{$\pm$  2.13}            & \textbf{74.28}        \tiny{$\pm$ 0.35 }        \\
ACD                            & 57.21             \tiny{$\pm$  4.00}            & \multicolumn{1}{c|}{\textbf{60.95} \tiny{$\pm$ 0.99}}           & 63.87          \tiny{$\pm$ 1.12}           & \textbf{70.09}        \tiny{$\pm$ 1.52}             & 30.83                \tiny{$\pm$ 0.53}        & \multicolumn{1}{c|}{\textbf{33.84}      \tiny{$\pm$ 1.04}}             & 37.19           \tiny{$\pm$ 0.82}            & \textbf{42.11}        \tiny{$\pm$ 2.29}        \\
ABSC                           & 84.06             \tiny{$\pm$  2.02}            & \multicolumn{1}{c|}{\textbf{87.58}       \tiny{$\pm$ 0.73 } }        & 86.98           \tiny{$\pm$ 0.67}            & \textbf{87.41}      \tiny{$\pm$ 0.98}      & 84.09                \tiny{$\pm$  1.40}        & \multicolumn{1}{c|}{\textbf{86.07}     \tiny{$\pm$ 0.20}}              & 85.16           \tiny{$\pm$ 2.81}            & \textbf{86.60}        \tiny{$\pm$ 0.38}        \\
COSC                           & 83.38  \tiny{$\pm$  0.62}  &\multicolumn{1}{c|}{ \textbf{84.95}\tiny{$\pm$ 0.83} }        & 83.78           \tiny{$\pm$ 0.99}            & \textbf{84.77}     \tiny{$\pm$ 1.11}       & \textbf{84.20}        \tiny{$\pm$ 0.30}        & \multicolumn{1}{c|}{\textbf{84.20}     \tiny{$\pm$ 1.85}}          & 85.78           \tiny{$\pm$ 0.39}            & \textbf{86.55}        \tiny{$\pm$  0.37}      \\
AOOE                           & 68.29             \tiny{$\pm$  5.04}           & \multicolumn{1}{c|}{\textbf{68.53}        \tiny{$\pm$ 2.52}}              & 75.72           \tiny{$\pm$ 1.31}            & \textbf{79.22}      \tiny{$\pm$ 0.90 }      & 65.58                \tiny{$\pm$ 1.02}        & \multicolumn{1}{c|}{\textbf{67.56}     \tiny{$\pm$ 1.76}}               & 74.82            \tiny{$\pm$ 0.87}            & \textbf{76.49}        \tiny{$\pm$ 1.26}        \\
\midrule
ASPE                           & 48.57             \tiny{$\pm$  4.59}            & \multicolumn{1}{c|}{\textbf{55.35}       \tiny{$\pm$ 1.82}}  & 60.31         \tiny{$\pm$ 1.66}            & \textbf{61.77}       \tiny{$\pm$ 2.22}      & 50.92                \tiny{$\pm$ 1.57}        & \multicolumn{1}{c|}{\textbf{58.79}        \tiny{$\pm$ 1.38}}           & 58.94          \tiny{$\pm$ 1.32}            & \textbf{63.11}       \tiny{$\pm$ 0.47}      \\
AOPE                           & 34.26             \tiny{$\pm$  0.89}            & \multicolumn{1}{c|}{\textbf{44.86}       \tiny{$\pm$ 1.12}}       & 47.31          \tiny{$\pm$ 1.40}            & \textbf{56.87}       \tiny{$\pm$ 1.29}       & 33.52               \tiny{$\pm$ 4.32}        & \multicolumn{1}{c|}{\textbf{47.27}       \tiny{$\pm$ 1.53}}            & 47.73           \tiny{$\pm$ 1.14}            & \textbf{52.80}        \tiny{$\pm$ 0.49}          \\
CSPE                           & 47.12             \tiny{$\pm$  3.27}            & \multicolumn{1}{c|}{\textbf{51.90}       \tiny{$\pm$ 2.23} }     & 57.63           \tiny{$\pm$ 1.80}            & \textbf{61.64}     \tiny{$\pm$ 1.41}        & 25.52                \tiny{$\pm$  2.07}        & \multicolumn{1}{c|}{\textbf{28.92}      \tiny{$\pm$ 1.43}}           & 31.85           \tiny{$\pm$ 0.69}            & \textbf{36.76}        \tiny{$\pm$ 1.33}          \\ \midrule

AOSTE                          & 34.20             \tiny{$\pm$  2.23}            & \multicolumn{1}{c|}{\textbf{44.96} \tiny{$\pm$ 1.55}}                & 46.81         \tiny{$\pm$ 3.46}            & \textbf{53.71}       \tiny{$\pm$ 0.54}        & 28.49                \tiny{$\pm$  9.70}        & \multicolumn{1}{c|}{\textbf{47.14}       \tiny{$\pm$ 3.96}}      & 46.36          \tiny{$\pm$  2.66}            & \textbf{52.83}       \tiny{$\pm$ 0.51 }         \\
ACSTE                          & 27.21             \tiny{$\pm$  2.74}            & \multicolumn{1}{c|}{\textbf{39.00} \tiny{$\pm$ 3.02}}             & 39.59          \tiny{$\pm$ 3.42}            & \textbf{47.74}       \tiny{$\pm$ 2.29}          & 19.31               \tiny{$\pm$ 1.31}        & \multicolumn{1}{c|}{\textbf{22.57}        \tiny{$\pm$ 0.68}}          & 25.91           \tiny{$\pm$ 0.63}            & \textbf{31.43}       \tiny{$\pm$ 0.99}               \\ \midrule
ACOSQE                         & 18.84             \tiny{$\pm$ 2.61}           & \multicolumn{1}{c|}{\textbf{26.73}  \tiny{$\pm$1.82}}               & 30.72           \tiny{$\pm$ 1.78}            & \textbf{36.75}        \tiny{$\pm$1.21}               & 14.05                \tiny{$\pm$ 0.60 }       & \multicolumn{1}{c|}{\textbf{16.00}        \tiny{$\pm$ 1.32}}               & 19.40           \tiny{$\pm$  0.68}            & \textbf{22.61}      \tiny{$\pm$  0.61 }           \\ \midrule
Ave.                           & {51.34}           & \multicolumn{1}{c|}{\textbf{57.29}}          & {60.10}           & {\textbf{64.58}}           & {45.50}           & \multicolumn{1}{c|}{\textbf{51.05}}          & {53.10}           & {\textbf{56.87}}           \\ \bottomrule
\end{tabular}%

}
\caption{The F1 scores ($\pm$ std) comparison of 11 tasks between single models (Single) and \textsc{Unified-ABSA} (Unify) on two quadruple datasets under 32-shot and 64-shot settings. }
\label{tab:low-resource-table}
\end{table*}

It is often time-consuming and laborious to annotate data. We consider the low-resource setting. Specifically, $K$-Shot of training (and development) instances from the quadruple data are randomly sampled, where the true few-shot setting~\citep{DBLP:conf/nips/PerezKC21} is observed. Then the training instances of other ABSA tasks are obtained from these quadruple samples via the data conversion.

The results under the 32-shot and 64-shot settings are shown in Table~\ref{tab:low-resource-table}. Our \UnifiedABSA significantly outperforms task-specific models across all datasets and tasks (about 3\% $\sim$ 6\%). We also find that two-element extraction and three-element extraction tasks (e.g., AOPE and AOSTE) have the highest performance improvement. In addition, an interesting finding is that with the help of prompt learning, the single model can achieve excellent performance (around 83-86\%) on sentiment classification tasks (i.e., AOSC and COSC) with only a minimal number of samples. \citet{seoh-etal-2021-open} also made similar observations, which performed ABSC and COSC tasks based on BERT~\citep{devlin-etal-2019-bert} and GPT-2~\citep{radford2019gpt2}. Despite its success, \UnifiedABSA still achieves substantial improvements over them. These results demonstrate our effectiveness.

\begin{figure}[t]
\centering
  \setlength{\belowcaptionskip}{-0.5cm}
  \setlength{\abovecaptionskip}{-0.3cm}
\begin{tabular}{p{7cm}p{7cm}}
\centering
\begin{minipage}{0.5\textwidth}

    \hbox{\hspace{0.05em}\includegraphics[width=0.75\columnwidth,height=0.15\textheight ]{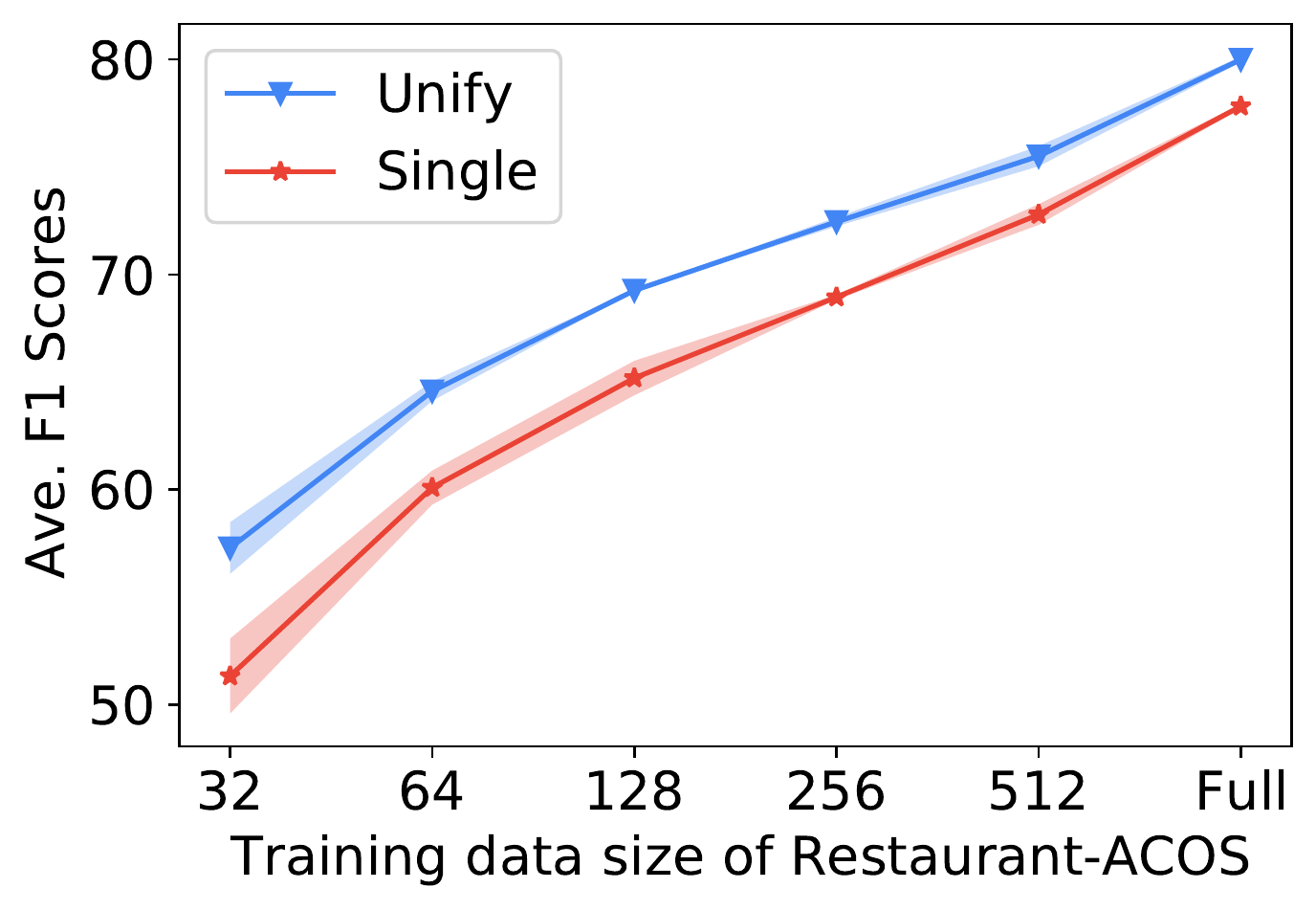}}
  \end{minipage}\vspace{0.3em}
\\
\begin{minipage}{0.5\textwidth}

    \hbox{\hspace{0.05em}\includegraphics[width=0.75\columnwidth,height=0.15\textheight ]{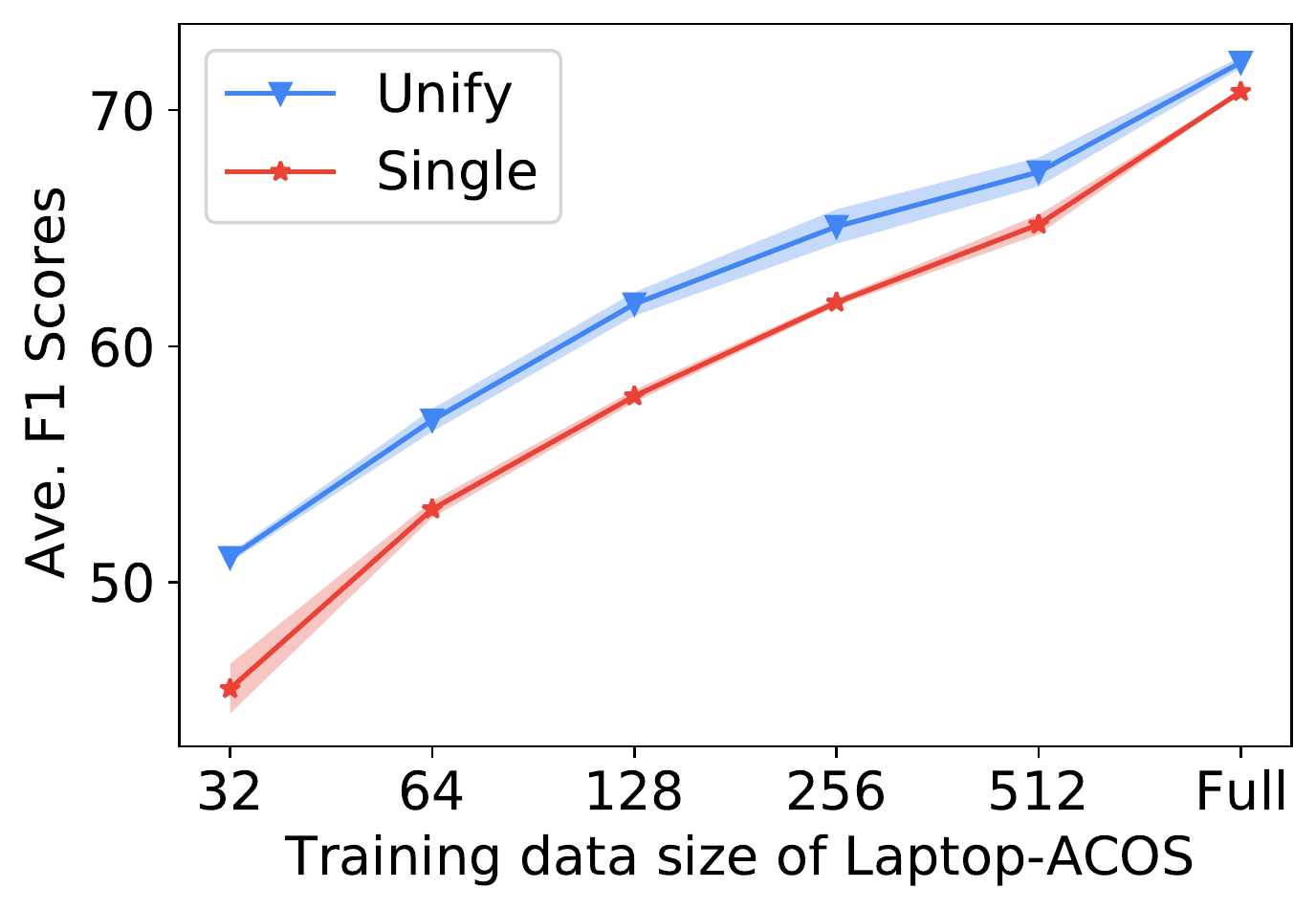}}
  \end{minipage}\vspace{0.3em}\\
\end{tabular}
\caption{The average performance ($\pm$ standard deviation) of 11 tasks comparison among single models (Single) and \UnifiedABSA (Unify) on two datasets under different training data size regimes.}
\label{fig:two-data-performance-curve}
\end{figure}

\noindent\textbf{Data Efficiency}. \quad In addition to the experiments on the above 32-shot and 64-shot settings, we then study the data efficiency. That is to say, \emph{does the model performance continue to improve as the amount of the data increases?} \emph{Furthermore, what is the relationship between the amount of data and the degree of performance improvement?} We conduct experiments in scenarios with different sizes of training data. As shown in Figure~\ref{fig:two-data-performance-curve}, \UnifiedABSA consistently outperforms single dedicated models under the different training data sizes on two datasets. The performance gap between them slowly decreases as the data size increases. Besides, we can observe that our \UnifiedABSA can achieve approximately the same performance with half of the data size as dedicated models when the data size is less than 512. These results indicate our advantages in terms of data efficiency.

\subsection{Effect of Components of \USI}

We compare several simplified versions of \UnifiedABSA to better understand the contribution of each component of \USI. Table~\ref{tab:ablation-on-usi} shows the results under the 32-shot and full data settings. There is the \emph{task name}, \emph{options} (\emph{sentiment options} and \emph{category options}), and \emph{template} in \USI, which can be considered as three distinguishers between tasks. 

\textbf{In the 32-shot setting,} we can observe that the task name is not necessary for Restaurant-ACOS (but a little boost after the removal), while it is indeed necessary for Laptop-ACOS (with 1.07\% performance improvements). Intuitively, the options and template boost the \UnifiedABSA by providing rich task-relevant information. Besides, it yields inferior performance when only the task name is utilized (i.e., w/o template \& options). Finally, the performance dramatically drops after the removal of \USI, indicating that it is difficult for a unified model to distinguish between different tasks and learn without instruction. \textbf{In the full-data setting,} each component of \USI contributes to performance. The relatively small performance gap between the first four variants and \UnifiedABSA can be considered as the combined effect of the powerful PLM and the full-data setting. Similarly to the 32-shot setting, once without \USI, the model could be confused, leading to a dramatic drop in performance.

\begin{table}[h]
\centering
\caption{Ablation study for \textsc{USI} on Restaurant-ACOS and Laptop-ACOS in terms of the average performance over 11 tasks. Note that ``options'' includes both sentiment options and category options.  ``tem.'' and ``opt.'' denote the template and options, respectively.}

\scalebox{0.90}{
\begin{tabular}{l|cc|cc}
\toprule
\multirow{2}{*}{\textbf{Model}} & \multicolumn{2}{c|}{\textbf{32-Shot}} & \multicolumn{2}{c}{\textbf{Full data}} \\  \cline{2-5}
                                & \textbf{Rest.}     & \textbf{Lap.}    & \textbf{Rest.}      & \textbf{Lap.}     \\ \midrule
w/o task name                   & \textbf{57.37}              & 49.98            & 79.91              & 71.84             \\
w/o options                     & 55.36              & 49.91            & 79.76              & 71.76             \\
w/o template                    & 55.14              & 48.62            & 79.54              & 71.92             \\
w/o tem. \& opt.                & 52.14              & 46.80            & 79.74              & 71.97             \\
w/o \textsc{USI}                         & 26.87              & 26.92            & 34.50              & 33.49             \\ \midrule
\textsc{UnifiedABSA}                     & 57.29              & \textbf{51.05}            & \textbf{79.99}    & \textbf{72.02}             \\ \bottomrule
\end{tabular}}
\label{tab:ablation-on-usi}
\end{table}

\subsection{Effect of Sampling Strategies}
\label{sec:ablation-on-sampling}

The data sampling strategy, i.e., how much data from each task should be sampled for training, plays a vitally important role during multi-task learning~\cite{raffel2020t5,DBLP:journals/corr/abs-1907-05019}. We conduct experiments to investigate the effect of different sampling strategies on Restaurant-ACOS and Laptop-ACOS under the 32-shot and full-data settings. We consider the following two strategies: random sampling (i.e., a training batch is randomly sampled from the mixture of all task samples) and uniform sampling (i.e., uniformly sampling training data for each task within a batch). \UnifiedABSA adopts the random sampling strategy by default. Besides, the uniform sampling strategy includes two counterparts, undersampling and oversampling. In undersampling, we sample training instances for all the tasks to the task with the lowest number of instances. Instead, we achieve the oversampling via random replication of instances to the task with the highest number of instances.

As shown in Table~\ref{tab:ablation-on-sampling}, we can observe that the over-sampling yields better performance than the undersampling since the latter has much less training data. Furthermore, the random sampling achieves highly competitive performance compared with the uniform sampling (the oversampling counterpart) across all cases. We report the results of random sampling for all main experiments. We leave the exploration for better sampling strategies to future work.

% \usepackage{multirow}
\begin{table}[h]
\centering
\caption{The average performance comparison between different sampling strategies within a batch on Restaurant-ACOS and Laptop-ACOS under the 32-shot and full data settings. \textsc{UnifiedABSA} adopts the random sampling strategy by default.}

\scalebox{0.90}{
\begin{tabular}{l|cc|cc}
\toprule
\multirow{2}{*}{\textbf{Sampling}} & \multicolumn{2}{c|}{\textbf{32 shot}}           & \multicolumn{2}{c}{\textbf{Full data}}          \\ \cline{2-5} 
                                   & \textbf{Rest.} & \textbf{Lap.} & \textbf{Rest.} & \textbf{Lap.} \\ \midrule
Uniform (under)                       & 52.91                    & 46.46                & 78.85                    & 71.05                \\
Uniform (over)                       & 56.89                    & \textbf{51.67}       & 79.73                    & \textbf{72.15}       \\ \midrule
Random              & \textbf{57.29}           & 51.05                & \textbf{79.99}           & 72.02                \\ \bottomrule
\end{tabular}}
\label{tab:ablation-on-sampling}
\end{table}

\subsection{Further Analysis}
\label{sec:further-analysis}

\begin{figure}[h]
\centering 
\includegraphics[width=0.9\columnwidth]{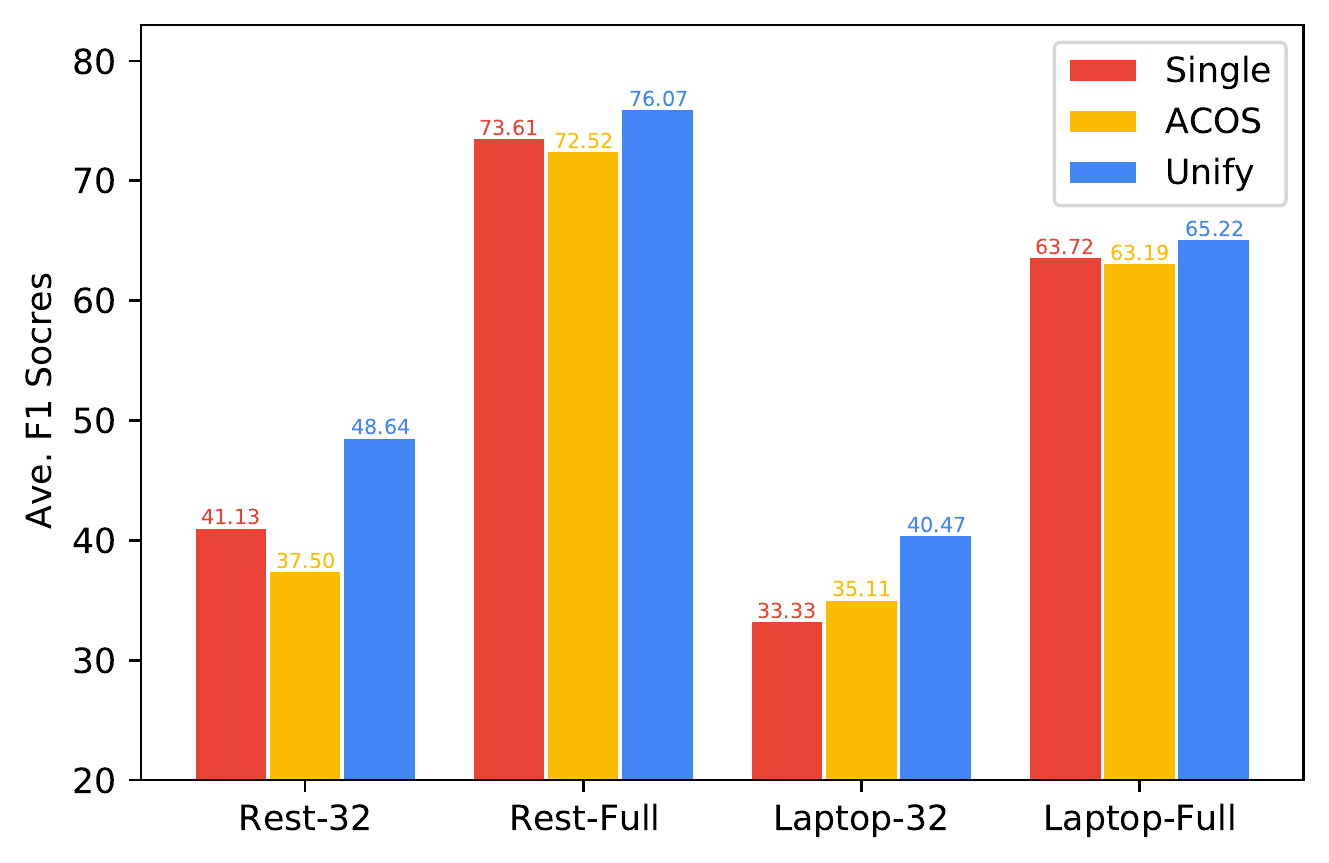} 
\caption{Average performance comparison among single models (Single), the ACOSQE-specific model (ACOS), and \UnifiedABSA (Unify) on Restaurant-ACOS and Laptop-ACOS under the 32-shot and full-data settings. Since the ACOS model is not readily adaptable to some tasks with a given aspect/category (i.e., AOSC, COSC, and AOOE), we report the average performance over the other eight tasks.} 
\label{fig:multi-task-baseline-performance-comparison}
\end{figure}

\noindent\textbf{Can Specialist for ACOSQE Be A Good Versatile ABSA Model?} \quad ACOSQE, as the most complex ABSA task so far, aims to extract sentiment quadruples. The prediction of (almost) all other ABSA tasks can be derived from its predicted results. Therefore, we consider the model dedicated to ACOSQE as a baseline\footnote{It is also trained based on the text-to-text \emph{Paraphrase}.} for the versatile ABSA model. Although the dedicated ACOSQE model can support almost all ABSA tasks, it is often suboptimal (yielding inferior performance to dedicated models, not to mention the \UnifiedABSA) due to the coupling of multiple sentiment elements, as shown in Figure~\ref{fig:multi-task-baseline-performance-comparison}. Nevertheless, we also find that it unexpectedly performed better than UnifiedABSA for category-related tasks such as ACD on Laptop-ACOS. Considering that the Laptop-ACOS has more categories (121 vs. 13) compared to Restaurant-ACOS, we conjecture that the modeling of the dedicated ACOSQE model provides more help for category-related tasks on the Laptop-ACOS. However, our \UnifiedABSA consistently outperforms the dedicated model for ACOSQE across datasets and data scenarios, which shows the superiority of our model.

\begin{table}[h]
\centering

\scalebox{0.90}{
\begin{tabular}{l|c}
\toprule
\textbf{Model}               & \textbf{\#Storage} \\ \midrule
Task-specialized Models            & 11$T$       \\
Multi-task Baselines & $T$         \\ \midrule
\textsc{UnifiedABSA}         & $T$        \\ \bottomrule
\end{tabular}}
\caption{Comparison of the storage space required by different models to support 11 ABSA tasks. $T$ denotes the storage space size occupied by T5.}
\label{tab:storage-efficiency}
\end{table}

\noindent\textbf{Storage Efficiency} \quad In real scenarios, it becomes impractical to separately train task-specific models for each downstream task, especially as the language model size increases, e.g., 42 GB for each copy of T5-XXL. As illustrated in Table~\ref{tab:storage-efficiency}, our \UnifiedABSA shows a more efficient way to build a general-purpose ABSA system. Specifically, \UnifiedABSA saves more than 90\% of storage space compared to dedicated models and yields superior performance both in fully-supervised (\S~\ref{sec:full-data-exp}) and low-resource settings (\S~\ref{sec:low-resource-exp}). Compared to some ABSA models that can support multiple tasks (e.g., the ACOSQE-specific model), our model can consistently perform better than them, even though they require the exact storage cost as our model.

\section{Conclusion}

In this work, we present \UnifiedABSA, a general-purpose ABSA framework based on multi-task instruction tuning which can uniformly model various ABSA tasks and take full advantage of the relationship between tasks. Moreover, it is easy to scale to new tasks without model architecture modifications. Compared to training the dedicated system on specific data for each task, \UnifiedABSA only requires the data from the ACOSQE task. It performs better with less storage space in the fully-supervised and low-resource setting. Further experiments exhibit that our model requires only half the data of the dedicated model to achieve comparable performance in low-resource scenarios. These empirical results demonstrate that \UnifiedABSA is scalable and more data- and storage-efficient, showing the potential for practical application. We hope our study could inspire more explorations for the general-purpose ABSA system.

\section*{Limitations}

Although our model has shown excellent performance and scalability, there are still some limitations for further improvement. 

Firstly, \UnifiedABSA is based on instructions described in natural language, which usually require much manual effort to design. To this end, soft prompts or instructions can be considered an alternative to automate prompt design, which can be learned from downstream task training data~\citep{gao-etal-2021lm-bff,lester-etal-2021-prompt-tuning}.

Secondly, is there a better unified ABSA framework? It should fulfill the following characteristics: 
(1) versatility, i.e., can handle multiple ABSA tasks; 
(2) quality, i.e., achieving superior performance; 
(3) scalability, i.e., easy to scale to new tasks with just minor or even no architecture modifications. Our work is only a preliminary exploration.

Thirdly, our study simplifies the application scenario. We only consider multi-task learning in a single domain in English. However, the actual scenario is the open-domain, multi-lingual, and multi-modal. We leave it to future work. 

Lastly, our framework only focuses on ABSA tasks. Intuitively, this framework can also be applied to other NLP tasks, such as \emph{information extraction}, via well-designed task instructions.

\section*{Ethics Statement}

We honor and support the ACL Ethics Policy. Our experiments are conducted on existing public standard datasets with no attached privacy or ethical issues. Our study does not need colossal GPU resources like PLMs. Our model is more data-efficient and storage-efficient compared to task-specific models.

% \section*{Acknowledgements}
% This document has been adapted by Yue Zhang, Ryan Cotterell and Lea Frermann from the style files used for earlier ACL and NAACL proceedings, including those for 
% ACL 2020 by Steven Bethard, Ryan Cotterell and Rui Yan,
% ACL 2019 by Douwe Kiela and Ivan Vuli\'{c},
% NAACL 2019 by Stephanie Lukin and Alla Roskovskaya, 
% ACL 2018 by Shay Cohen, Kevin Gimpel, and Wei Lu, 
% NAACL 2018 by Margaret Mitchell and Stephanie Lukin,
% Bib\TeX{} suggestions for (NA)ACL 2017/2018 from Jason Eisner,
% ACL 2017 by Dan Gildea and Min-Yen Kan, NAACL 2017 by Margaret Mitchell, 
% ACL 2012 by Maggie Li and Michael White, 
% ACL 2010 by Jing-Shin Chang and Philipp Koehn, 
% ACL 2008 by Johanna D. Moore, Simone Teufel, James Allan, and Sadaoki Furui, 
% ACL 2005 by Hwee Tou Ng and Kemal Oflazer, 
% ACL 2002 by Eugene Charniak and Dekang Lin, 
% and earlier ACL and EACL formats written by several people, including
% John Chen, Henry S. Thompson and Donald Walker.
% Additional elements were taken from the formatting instructions of the \emph{International Joint Conference on Artificial Intelligence} and the \emph{Conference on Computer Vision and Pattern Recognition}.


\bibliography{emnlp2022}
\bibliographystyle{acl_natbib}

\appendix

% \clearpage

\section{The Detail of Conversion from ACOS Quadruple Data to Other Tasks}
\label{sec-appendix:acos-data-conversion}

Generally, we extract the simple sentiment tuples from the quadruples and then de-duplicate them to get the labels corresponding to simple sentiment tuples. Suppose the original ACOS data has $n$ items, and since one sample may correspond to more than a quadruple, more than $n$ items will be obtained when degraded to other tasks. Since there are implicit aspect terms and opinion terms in the ACOS data~\citep{cai-etal-2021-acos}, we need to follow certain principles to handle this situation. Specifically, we consider the implicit aspect term (opinion term) if and only if there is a corresponding aspect category (sentiment).

\section{Experiments on Task Complementarity}
\label{sec-appendix:effect-of-training-tasks}
We study the effect of different tasks on multi-task learning. For simplicity, we group 11 ABSA tasks into  4 task categories based on whether they involved the four sentiment elements (i.e., aspect, category, opinion, sentiment). This often means that there are a few tasks that fall into several task categories at the same time (e.g., ACOSQE). We then train jointly for each task category and compare the performance with \UnifiedABSA on the tasks within the task category. Globally, all ABSA tasks can gain from each other, as shown in Table~\ref{tab:ablation-on-tasks}. Specifically, \UnifiedABSA still achieves the best results on more than half of the tasks. We can also find that the model trained jointly only on the aspect category-related tasks achieves the best results on CSPE and ACD. However, the best results are not achieved on the other category related tasks (i.e., ACSTE and ACOSQE), which indicates that other tasks also contribute to their performance. The above results show the effectiveness of \UnifiedABSA, jointly training all ABSA tasks, and also suggest much room for improvement in multi-task learning.

% \usepackage{multirow}
\begin{table*}[]
\centering
\begin{tabular}{l|ccccc|ccccc}
\toprule
\multirow{2}{*}{\textbf{Task}} &
  \multicolumn{5}{c|}{\textbf{Restaurant-ACOS}} &
  \multicolumn{5}{c}{\textbf{Laptop-ACOS}} \\ \cline{2-11} 
 &
  \textbf{aspect} &
  \textbf{cate.} &
  \textbf{opinion} &
  \multicolumn{1}{c|}{\textbf{senti.}} &
  \textbf{all} &
  \textbf{aspect} &
  \textbf{cate.} &
  \textbf{opinion} &
  \multicolumn{1}{c|}{\textbf{senti.}} &
  \textbf{all} \\ \midrule

ATE &
  63.50 &
  \textbackslash{} &
  \textbackslash{} &
  \multicolumn{1}{c|}{\textbackslash{}} &
  \textbf{65.36} &
  \textbf{69.79} &
  \textbackslash{} &
  \textbackslash{} &
  \multicolumn{1}{c|}{\textbackslash{}} &
  69.22 \\
ACD &
  \textbackslash{} &
  \textbf{61.91} &
  \textbackslash{} &
  \multicolumn{1}{c|}{\textbackslash{}} &
  60.95 &
  \textbackslash{} &
  \textbf{34.48} &
  \textbackslash{} &
  \multicolumn{1}{c|}{\textbackslash{}} &
  33.84 \\
AOOE &
  \textbackslash{} &
  \textbackslash{} &
  64.43 &
  \multicolumn{1}{c|}{\textbackslash{}} &
  \textbf{68.53} &
  \textbackslash{} &
  \textbackslash{} &
  64.01 &
  \multicolumn{1}{c|}{\textbackslash{}} &
  \textbf{67.56} \\ 

ABSC &
  \textbackslash{} &
  \textbackslash{} &
  \textbackslash{} &
  \multicolumn{1}{c|}{86.22} &
  \textbf{87.58} &
  \textbackslash{} &
  \textbackslash{} &
  \textbackslash{} &
  \multicolumn{1}{c|}{85.29} &
  \textbf{86.07} \\
COSC &
  \textbackslash{} &
  \textbackslash{} &
  \textbackslash{} &
  \multicolumn{1}{c|}{84.72} &
  \textbf{84.95} &
  \textbackslash{} &
  \textbackslash{} &
  \textbackslash{} &
  \multicolumn{1}{c|}{\textbf{85.14}} &
  84.20 \\
  
\midrule

ASPE &
  52.02 &
  \textbackslash{} &
  \textbackslash{} &
  \multicolumn{1}{c|}{54.10} &
  \textbf{55.35} &
  58.21 &
  \textbackslash{} &
  \textbackslash{} &
  \multicolumn{1}{c|}{56.81} &
  \textbf{58.79} \\
AOPE &
  40.44 &
  \textbackslash{} &
  \textbf{45.42} &
  \multicolumn{1}{c|}{\textbackslash{}} &
  44.86 &
  43.49 &
  \textbackslash{} &
  46.31 &
  \multicolumn{1}{c|}{\textbackslash{}} &
  \textbf{47.27} \\
CSPE &
  \textbackslash{} &
  \textbf{53.90} &
  \textbackslash{} &
  \multicolumn{1}{c|}{53.52} &
  51.90 &
  \textbackslash{} &
  \textbf{30.01} &
  \textbackslash{} &
  \multicolumn{1}{c|}{28.27} &
  28.92 \\ \midrule

AOSTE &
  40.96 &
  \textbackslash{} &
  42.15 &
  \multicolumn{1}{c|}{44.48} &
  \textbf{44.96} &
  44.09 &
  \textbackslash{} &
  41.88 &
  \multicolumn{1}{c|}{42.38} &
  \textbf{47.14} \\
ACSTE &
  34.52 &
  37.88 &
  \textbackslash{} &
  \multicolumn{1}{c|}{\textbf{39.06}} &
  39.00 &
  \textbf{24.25} &
  23.06 &
  \textbackslash{} &
  \multicolumn{1}{c|}{21.93} &
  22.57 \\ \midrule
  
ACOSQE &
  22.45 &
  23.87 &
  20.99 &
  \multicolumn{1}{c|}{26.46} &
  \textbf{26.73} &
  14.81 &
  13.73 &
  13.16 &
  \multicolumn{1}{c|}{14.97} &
  \textbf{16.00} \\

  \bottomrule
\end{tabular}
\caption{Performance comparison between partial tasks for unified training and \textsc{Unified-ABSA} under the 32-shot setting. \emph{aspect}, \emph{cate.}, \emph{opinion}, and \emph{senti.} denotes that we only consider tasks involving aspect, category, opinion, and sentiment in the training. \emph{all} means that all tasks participate in training, i.e., \textsc{unifiedABSA}.}
\label{tab:ablation-on-tasks}
\end{table*}

\section{Case Study}
Figure~\ref{tab:case-study} compares the predicted results between single models and \UnifiedABSA on 11 tasks. It can be observed that single models consistently predict aspect term ``\emph{kafta plate}'' as ``\emph{plate}'', resulting in all wrong predictions on the aspect-related tasks. However, it can be successfully predicted by \UnifiedABSA, while \UnifiedABSA also predicts one more wrong aspect category (restaurant general) and classifies it as positive sentiment.

\begin{table*}[ht]
\centering
\caption{The predicted results comparison between single models and \UnifiedABSA for the example ``\emph{I had the kafta plate, and it was perfect.}''. ``$\rightarrow$'' denotes performing opinion extraction or sentiment classification prediction based on a given aspect or category. The ``POS'' denotes the positive sentiment polarity.}
\renewcommand{\arraystretch}{1.2}
\scalebox{0.90}{
\begin{tabular}{l|m{5.5cm}<{\centering}|m{6.5cm}<{\centering}}
\toprule
\textbf{Task} & \textbf{\emph{Single}}               & \textbf{\textsc{UnifiedABSA}}    \\ \midrule
ATE           & plate \ding{55}                               & kafta plate  \ding{51} \\ \hline
ACD           & food quality  \ding{51}                       & food quality  \ding{51} \\ \hline
AOOE          & kafta plate $\rightarrow$ perfect  \ding{51}  & kafta plate $\rightarrow$ perfect  \ding{51}                                                                 \\ \hline
ABSC          & kafta plate $\rightarrow$ POS  \ding{51}      & kafta plate $\rightarrow$ POS  \ding{51}                                                                     \\ \hline
COSC          & food quality $\rightarrow$ POS  \ding{51}     & food quality $\rightarrow$ POS  \ding{51}                                                                    \\ \hline
AOPE          & (plate, perfect) \ding{55}                    & (kafta plate, perfect)  \ding{51}                                                                            \\ \hline
ASPE          & (plate, POS) \ding{55}                        & (kafta plate, POS)  \ding{51}                                                                                \\ \hline
\multirow{2}{*}{CSPE} & 
\multirow{2}{*}{(food quality, POS)  \ding{51}}&  (food quality, POS)   \ding{51} \\ 
& & (restaurant general, POS)  \ding{55} \\ \hline
ACSTE         & (plate, food quality, POS) \ding{55}          & (kafta plate, food quality, POS)  \ding{51}                                                                  \\ \hline
AOSTE         & (plate, perfect, POS) \ding{55}               & (kafta plate, perfect, POS)  \ding{51}                                                                       \\ \hline
ACOSQE        & (plate, food quality, perfect, POS) \ding{55} & (kafta plate, food quality, perfect, POS)  \ding{51}                                                         \\ \bottomrule
\end{tabular}}
\label{tab:case-study}
\end{table*}

\end{document}